\documentclass{article}
\usepackage{ijcai22}

\usepackage{times}
\usepackage{amssymb}
\usepackage{multirow}
\usepackage{graphicx}
\usepackage{arydshln}
\usepackage[section]{placeins}
\usepackage{booktabs}

\title{A Mask-Based Adversarial Defense Scheme}

\author{
    Author Name
    \affiliations
    Affiliation
}

\author{
Weizhen Xu
\and
Chenyi Zhang\and
Fangzhen Zhao\And
Liangda Fang
\affiliations
Jinan University
}

\begin{document}

\maketitle

%\begin{frontmatter}



%\title{MAD:A mask-based adversarial defense training method}


%\author[inst1]{Author One}

\begin{abstract}
%A mask-based adversarial defense training method(MAD) is proposed in this paper. In the proposed method, parts of the images are masked to reduce the adversarial disturbance. Compared with other adversarial defense methods, the proposed method does not need additional network structure or loss function. Experimental results on different data sets and networks show that the proposed method can effectively improve the adversarial defense ability of the network when facing most of the adversarial attacks methods.

Adversarial attacks hamper the functionality and accuracy of Deep Neural Networks (DNNs) by meddling with subtle perturbations to their inputs. %which poses a long-standing challenge. 
In this work, we propose a new Mask-based Adversarial Defense scheme (MAD) for DNNs to mitigate the negative effect from adversarial attacks. %Our method promotes the robustness of a DNN by adding fuzzy elements at its training stage, i.e., we randomly mask a portion of the provided training images to make the DNN more tolerant to small input perturbation. 
To be precise, our method promotes the robustness of a DNN by randomly masking a portion of potential adversarial images, and as a result, the %classification result 
output of the DNN becomes more tolerant to minor input perturbations.
Compared with existing adversarial defense techniques, our method does not need any additional denoising structure, nor any change to a DNN's design. We have tested this approach on a collection of DNN models for a variety of data sets, and the experimental results confirm that the proposed method can effectively improve the defense abilities of the DNNs against all of the tested adversarial attack methods. In certain scenarios, the DNN models trained with MAD have improved classification accuracy by as much as %$20\%$ to 
$90\%$ compared to the original models that are given adversarial inputs.
\end{abstract}

%\begin{keyword}
%adversarial defense \sep neural network
%\end{keyword}

%\end{frontmatter}



\section{Introduction}\label{sec:introduction}

Deep Neural Networks (DNNs) have achieved great success in the past decade in research areas such as image classification, natural language processing, and data analytics, with a variety of application domains like banking, financial services and insurance, IT \& Telecom, manufacturing, and healthcare etc.,~\cite{AMR}. However, %more recently, 
researchers have discovered that it is possible to introduce human imperceptible perturbations to inputs of a DNN in a certain way that induces incorrect or misleading outputs from the DNN at the choice of an adversary~\cite{szegedy2014,goodfellow2015,papernot2016}.
%With the application of neural network in automatic driving, medical and other fields, higher requirements are put forward for neural network. Some adversarial attack methods aim to get the legal input that makes the neural network make wrong decisions. It is important for the network to make correct decisions when facing adversarial examples generated by these adversarial attacks methods. 
%To counter adversarial attacks, t
As of today, the bulk of the existing
defense approaches can be roughly grouped in two %directions, 
categories, i.e., they either focus on the detection of adversarial inputs (e.g.,~\cite{LID,InfluenceFunction,FeatureSqueezing}), or to take steps to strengthen DNNs (e.g.,~\cite{meng2017magnet,madry2018towards,mustafa2019adversarial}), making them more robust to withstand perturbations on inputs.

\begin{figure}%[]
    \centering
    \includegraphics[width=\linewidth]{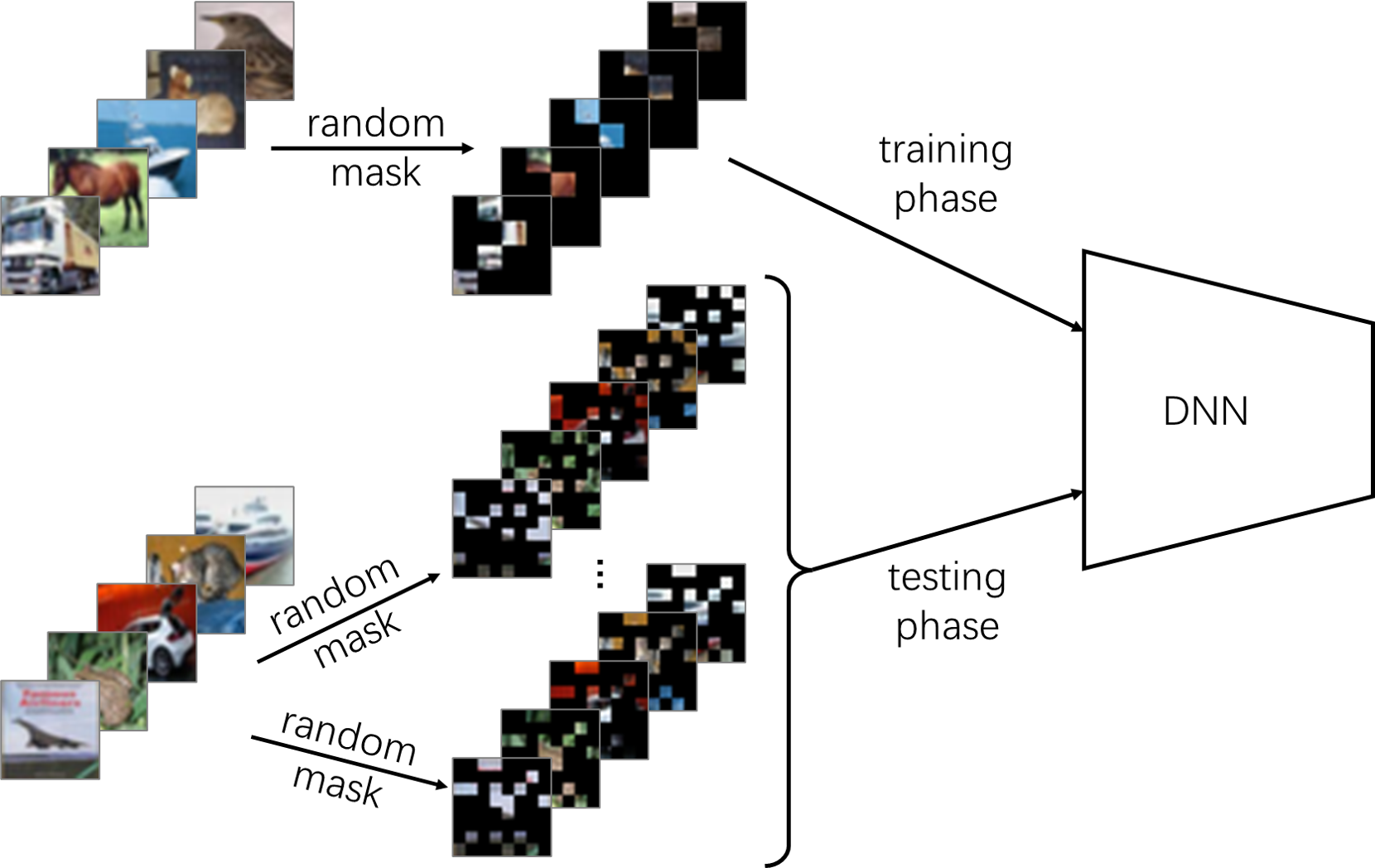}
    \caption{An overview of the MAD scheme}
    \label{fig:process}
\end{figure}

In this paper, we follow the latter path by enhancing  robustness in the decision procedure of DNNs. Again, there exist a rich class of works in the school of adversarial defense. Some early works propose adversarial training which applies adversarial inputs together with clean inputs at the time of training, in order to reduce the effectiveness of adversarial samples on DNNs~\cite{madry2018towards,LeeLY20},  or distillation~\cite{PapernotM0JS16} which trains a model to reduce the magnitude of gradients against adversarial attacks. Other approaches boost robustness of DNNs by redesigning training methods~\cite{mustafa2019adversarial}, %or loss functions~\cite{PangXDD0Z20}, %Other recent works 
or apply %various 
filtering mechanisms that ``correct'' the adversarial inputs through autoencoder-based structures~\cite{meng2017magnet,salman2020denoised}. Motivated by a recent paper~\cite{he2021masked} which %applies an autoencoder to 
restores missing pixels %from masked random patches 
of an image, we devise a new adversarial defense scheme called a Mask-based Adversarial Defense training method (MAD) for DNNs that classify images. In addition, we believe that such a mechanism may also be applicable to improve DNN robustness in other scenarios.
Firstly, we have the following observations.
\begin{itemize}
    \item Information density is often low in images, as missing patches from an image can often be %easily
    mostly recovered by a neural network structure~\cite{he2021masked} or by a human brain (as you look at something through a fence). 
    \item Adversarial attacks on images usually introduce a minor perturbation to inputs, which may be reduced or cancelled by (randomly) covering part of the image.
\end{itemize}
%% our approaches
In this approach, we split an image into grids of pre-defined size (e.g., $4\times 4$), and randomly mask each grid with a given probability (e.g., $75\%$) into a default value (e.g., the black value in RGB for those masked pixels) to generate training samples. After the training, we also apply masking to images at the test stage (for the classification task), as illustrated in Fig.~\ref{fig:process}. Given an (unmasked) image, the test process is to be repeated a number of times, which makes a better chance of filtering out malicious perturbations, and the final decision is then defined as the mostly voted class taking into account the outputs for all randomly masked inputs. 

%Starting with the principle of adversarial attack, the reason for the success of adversarial attack is analyzed. In order to reduce the impact of adversarial disturbance on neural network, a \textbf{M}ask-based \textbf{A}dversarial \textbf{D}efense training method(MAD) is proposed in this paper, which enables the network maintain high classification accuracy when facing adversarial examples. The proposed method uses masked images for training and testing to improve the adversarial defense ability of the network. This means that the structure and loss function of the network do not need to be changed, so it has better generalization. Experimental results on different networks and data sets show that the proposed method can effectively improve the defense ability of neural network when facing most of the adversarial attacks methods.

\paragraph{Contribution.} In this work, we introduce a new %Mask-based Adversarial Defense training 
adversarial defence method MAD which applies masking at both the training phase and the test phase. This approach enjoys the following  properties.
\begin{itemize} %% to consider in the discussion
    \item Our method seems easily applicable to many existing DNNs. Compared with other adversarial defense methods, we do not need special treatment to the structures of DNNs, redesign of loss functions, or any (autoencoder-based) input filtering.
    \item The randomized masking at both the training phase and the test phase yields our method more resistant to adaptive white-box adversarial attacks.
    \item The experiment on a variety of models (LeNet, VGG and ResNet) has shown the effectiveness of our methods with a significant improvement in defense (up to $93\%$ precision gain) when facing various adversarial samples, compared to the models without MAD.  
\end{itemize}

\section{Related Work}\label{sec:relative}
%% Chenyi:
%% adversarial attack & defense --- focus on defense mechanisms: AE-based, AID-Purifier, ICCV2019, NIPS2021 etc.
%% self-supervised learning, including the mask-based approach

Since we work on an adversarial defense scheme, we conduct a brief overview to the mechanism of adversarial attacks and various types of adversarial defense works.

\subsection{Adversarial attack}\label{sec:relative_attack}

In general, an adversarial attack method generates tiny perturbations that are %subsequently 
applied to clean inputs, making them incorrectly classified by a DNN. Some of the earliest methods include box-constrainted L-BFGS~\cite{szegedy2014}, Fast gradient sign method (FGSM)~\cite{goodfellow2015} and Jacobian-based Saliency Map Attack (JSMA)~\cite{PapernotM0JS16}. Perhaps the most widely used attack method is FGSM, which is also included in our experiment. FGSM generates a perturbation for an image by computing the following. % value.
%The adversarial attack method generates tiny disturbances and adds them to the images, so that the images that were originally correctly classified are incorrectly classified by the neural network now. Fast gradient sign method (FGSM) \cite{goodfellow2015} obtains such a disturbance by calculating the gradient and increasing the loss of the input image, which is expressed as 
\begin{equation}
\delta = \epsilon \ sign(\triangledown_x L(\theta, x, y)) \label{equ:FGSM}
\end{equation}
where $L(\cdot)$ is the loss function used for neural network training which calculates the difference %when inputting the 
between label $y$ and the output produced from input $x$ for the DNN with parameter $\theta$. The direction of movement is obtained by the gradient of loss and constant $\epsilon$ which %control the moving distance so that the disturbance is tiny enough.  
restricts the norm of perturbation.
Basic iterative method (BIM) \cite{2016Adversarial} and projected gradient descent (PGD) \cite{madry2018towards} are the iteration and extension of FGSM to obtain better attack effect, respectively. %respectively expressed as 
%\begin{equation}
%x^{t+1} = Clip_{x, \epsilon}(x^t + \alpha \ %sign(\triangledown_x L(\theta, x, y))) \label{equ:BIM}
%\end{equation}
%\begin{equation}
%x^{t+1} = \Pi_{x+S}(x^t + \alpha \ sign(\triangledown_x L(\theta, x, y))) \label{equ:PGD}
%\end{equation}

%Chenyi: I comment out the following as we should focus more on adversarial defense than adversarial attacks. We only need to mention those attack methods used in our experiments.

\newcommand{\commentout}[1]{}
\commentout{
In the classification task, the cross entropy loss function is generally used, which is expressed as 
\begin{equation}
L=-\sum^M_{c=1}y_c\ log(p_c) \label{equ:CS}
\end{equation}
where $p_c$ is the probability of predicting the input as category $c$. $y_c$ is 1 when the input category is $c$, otherwise it is 0. Adversarial attack method mentioned above increase the value of loss, which means the prediction probability of the real category of the input needs to be reduced. When the prediction probability of the real category is lower than that of some other category, the image will be classified incorrectly, as the category with the largest prediction probability will be regarded as the result of classification. 

Therefore, if the impact of adversarial disturbance on the classification probability of the correct class can be reduced, that is, the reduction of the classification probability of the real class is relatively small, so that the reduced classification probability of the correct class is still greater than that of other classes, then the image can still be classified correctly.

The above adversarial attack method belongs to white box attack method, which needs to know the network parameters. Another kind of adversarial attack method, such as CW \cite{Carlini2017Towards} and DeepFool \cite{moosavi2016deepfool}, can realize black box attack. These methods finally realize the adversarial attack by adding adversarial disturbance to the image. This type of attack is often more difficult to defend, but the computational cost of the attack is also relatively high.
}%%% END of commentout

The other advanced adversarial attack methods used in our experiments are DeepFool~\cite{moosavi2016deepfool} and CW~\cite{Carlini2017Towards}. %Given an image, 
DeepFool computes an approximation of a minimal perturbation according to the DNN model's decision boundary. In detail, given an image, DeepFool iteratively computes a small perturbation, taking into account the linearized boundary of region that contains the current perturbed image, %and it is 
thus it can be shown that DeepFool is able to compute a smaller perturbation than FGSM while keeping similar successful attack ratios. Carlini and Wagner introduce adversarial attack methods that make quasi-imperceptible perturbations by restricting $\ell_0$, $\ell_2$ and $\ell_\infty$ norms. The CW attack not only breaks some of the well accepted adversarial defense methods such as defensive distillation~\cite{papernot2016}, but also achieves remarkable transferability, in the sense that it can also be used for \emph{black box} attacks when the parameters of a model is unknown.

\subsection{Adversarial defense}
\label{sec:relative_defense}

%In order to counter the adversarial attack method, many adversarial defense methods have been proposed so that the neural network can correctly classify the adversarial samples.
%The purpose of 
Adversarial defense aims to counter the adversarial effect and to make DNNs achieve performance on adversarial samples close to results on clean samples. Existing approaches can be roughly classified as (1) use of additional denoising structures before the DNN, and (2) enforcing the DNN to become more robust against adversarial attacks.
%, and (3) a combination of the above two methods.

%%filtering based methods:
%% MagNet, Dnoised smoothing, DefenseGAN, PixelDefend
\paragraph{Additional denoising structures.} MagNet~\cite{meng2017magnet} is a framework consisting of a detector that rejects adversarial samples that are far away the normal input manifolds and a reformer that finds the closest normal input if the adversarial input is not far from the manifolds. A similar approach is HGD~\cite{LiaoLDPH018} which applies high-level representation guided denoiser, so that it can be designed as a defense model that transforms adversarial inputs to easy-to-classify inputs.
Defense-GAN~\cite{samangouei2018defense} tries to model the distribution of %undisturbed
clean images, and it can find a close output to an adversarial image which does not contain the adversarial changes. Denoised smoothing~\cite{salman2020denoised} prepends a custom-trained denoiser to a DNN and uses randomized smoothing for training the combination which enforces a non-linear Lipschitz
property. PixelDefend~\cite{song2018pixeldefend} approximates the training distribution using a PixelCNN model and purifies images towards higher probability areas of the training distribution. Compared with our method, all these approaches introduce additional network structures which help to remove adversarial noise from inputs at the cost of increasing the number of parameters of the working model. In our experiment, we also empirically show that MAD outperforms MagNet and Denoised smoothing in most of the test cases.

%Some adversarial defense methods, such as MagNet \cite{meng2017magnet}, Defense-GAN\cite{samangouei2018defense}, Pixeldefend\cite{song2018pixeldefend} and denoised smoothing\cite{salman2020denoised}, use additional network structure to process the image before inputting them into the neural network to reduce the adversarial disturbance. The additional network structure increases the parameters of the network. 

\paragraph{DNN model Enhancement.} In this category, adversarial training~\cite{madry2018towards,LeeLY20} and defensive distillation~\cite{papernot2016} are among the early successful approaches. In particular, adversarial training introduces adversarial samples at the training stage to enhance resistance against attacks. Defensive distillation transfers one DNN to another with the same functionality but less sensitive to input perturbations. %Some other adversarial defense methods, realize adversarial defense by modifying the network.
However, both adversarial training and defensive distillation require substantially more training cost.
Introducing randomized variation to inputs and network parameters at the time of training as been discussed in~\cite{potdevin19,GuR14}.
Parseval networks \cite{pmlr-v70-cisse17a} limits the Lipschitz constant of linear, convolutional and aggregation layers in the network to be smaller than $1$, making these layers tight frames. PCL \cite{mustafa2019adversarial} separates the hidden feature of different categories by designing a new loss function and bypass network, so as to increase the difficulty of adversarial attacks and achieve the effect of adversarial defense. Since the existing DNN models are already being optimized regarding performance (e.g., neural network search (NAS) \cite{zoph2016neural} technology is widely used to find neural networks structure with better performance), the modification of DNN structures may introduce human bias, resulting in unpredictable performance degradation. In contrast, our approach MAD does not require any change to network structure or loss function design.

%\section{Method}
\section{Mask-based adversarial defense}
\label{sec:method}

\newcommand{\IN}{\mathcal{X}}
\newcommand{\OUT}{\mathcal{Y}}
\newcommand{\R}{\mathbb{R}}

In this paper we focus on DNNs that classify images. Formally, a DNN is a function $f_\theta$ which maps $\IN\subseteq\R^d$ to $\OUT$, where $\theta$ represents values for the network parameters that are determined at training, $d$ is the dimension of the input space, $\IN$ is the input space for images and $\OUT$ is a finite set of classes or labels to be returned from $f_\theta$. An adversarial input can be written as $x'=x+\delta$ with $x$ a clean input and $||\delta||_p<\varepsilon$, such that $f_\theta(x)\neq f_\theta(x')$, where $||\delta||_p$ is the $p$ norm of the perturbation $\delta$ and $\varepsilon>0$ is of negligible size, making $x$ and $x'$ in the same class to a human being.

%% Chenyi: need to find a place for the following
%Some research, like AID-Purifier\cite{hwang2021aid} and SOAP\cite{shi2020online}, generate disturbances opposite to adversarial disturbances and add them to images to reduce adversarial disturbances.%

%In order to make the neural network classify the adversarial samples correctly, the adversarial disturbance in the adversarial samples should be reduced, so as to weaken the effect of adversarial attack. As mentioned in section \ref{sec:relative_attack} and shown in the experiment in section \ref{sec:ex_disturbances}, when the adversarial disturbance is small enough to make the classification confidence of the correct category drop sufficiently small, the image can still be classified correctly. In the proposed method, a simpler way is adopted to reduce the adversarial disturbance: directly mask part of the adversarial disturbances. Since the adversarial disturbances part and the normal part in the images cannot be distinguished, both of them will be mask in the masked area. Some masked images with different mask rates are shown in figure \ref{fig:rate}. Some research, like Masked Autoencoders (MAE) \cite{he2021masked}, show that neural network can reconstruct the image with high quality using only some parts of the image. The classification task is easier than the reconstruction task. In addition, there is some redundant information in the image. Therefore, neural network can still classify partially masked images.

\subsection{Training a classifier for masked images}

A natural approach to counter an adversarial input $x'$ is to reduce the effect of perturbation $\delta$. %either by adding additional structures or by making a DNN less sensitive to perturbations. 
Given $\delta$ combined with the pixels of a natural image $x$, masking part of $x'$ would potentially reduce the adversarial effect of $\delta$. Fortunately, such a masking operation which ``covers'' parts of an image does not necessarily make the classification job more difficult for DNNs. Since natural images are heavy in spatial information redundancy, as shown in~\cite{he2021masked}, in which missing pixels of an image can be reconstructed by the state-of-the-art MAE model even though a large amount of input pixels are masked.\footnote{Our human brains possess similar power of recognizing (or even reconstructing) a partially masked object.} Since the task of classification is supposed to be easier than reconstruction, this leads us to our first step of training a DNN classifier for masked images, keeping the same network structure as the original DNN to be defended.

Define $\tau:\R^d\times C\rightarrow\R^d$ a masking operator, where $C$ is a pre-defined set which has its cardinality depending on %size of an input, size of a patch and the percentage of an input image to be covered.
the number of grids that cover the input.
%For example, masking $50\%$ of an $4\times 4$ image by using $2\times 2$ patches gives ${4 \choose 2} = 6$ possible ways of masking, so that in this case one may set $|C| = 6$. 
For example, masking a $4\times 4$ image by using $2\times 2$ grids gives $2^4 = 16$ possible ways of masking ($2$ possibilities for each grid), so that in this case one may set $|C| = 16$. (The extreme case is that all grids are masked, but this can hardly happen as we are to explain in our experiment.) 
Suppose the original classifier $f_\theta$ is trained with sample set $Z$, we use sample set $Z\times C$ to retrain the structure into a new DNN model $f_{\theta'}:\R^d\rightarrow\OUT$ that is used to ``simulate'' the performance of $f_\theta$, in the sense that given $x\in\IN$, the output of $f_{\theta'}(\tau(x,c))$ should approximate $f_\theta(x)$, given $c\in C$ randomly picked.

In the proposed method, images are partially masked in both the training and test phases. For our experiment, in the training phase, each image from the training samples of CIFAR-10~\cite{cifar10}  and SVHN~\cite{svhn} is divided into $8\times 8$ grids, and images from MNIST~\cite{mnist} are divided into $7\times 7$ grids, so that the length of a grid always divides the length of an image (CIFAR-10 and SVHN data sets contain $32\times 32$ images and MNIST data set contains $28\times 28$ images). 

\begin{table}[t]
    \centering
    \begin{tabular}{ccccc}
    \hline
    $\#$ tests    &  $1$   & $3$ & $5$ & $7$ \\
    \hline
    precision     &  $75.89\%$  &  $81.78\%$   & $82.65\%$ & $82.97\%$ \\
    \hline
    \end{tabular}
    \caption{The precision of retrained model for $3/4$ masked images from CIFAR-10 with multiple tests}
    \label{tab:preliminary}
\end{table}

We then conduct a preliminary study on the retrained  VGG16~\cite{VGG} classifier for masked images from CIFAR-10, the results shown in Table~\ref{tab:preliminary}. The original VGG16 model (i.e., $f_\theta$) has $84.18\%$ accuracy on the test data. The retrained model (i.e., $f_{\theta'}$) is obtained %from %training samples with $3/4$ pixels randomly masked by
by letting each $8\times 8$ grid have $3/4$ probability to be masked, and similarly, test images also have $3/4$ probability for %$3/4$ pixels randomly masked by 
each $4\times 4$ grid to be masked. Note that in Table~\ref{tab:preliminary}, ``$\#$tests'' indicates that we also combine results from multiple tests (i.e., $3$, $5$, or $7$ tests) of the same image with randomized masking applied, and the mostly occurred class of all test results is returned as output for that image. It can be easily seen that applying masking on images with a single test does imply performance degradation ($75.89\%$ dropped from $84.18\%$). However, it seems that the loss of precision can be remedied by applying randomized multiple tests on the same image followed by taking the most favored class, as one may see that in the case of combining $7$ tests a precision of $82.97\%$ can be achieved.

\begin{figure}[t]
    \centering
    \includegraphics[width=\linewidth]{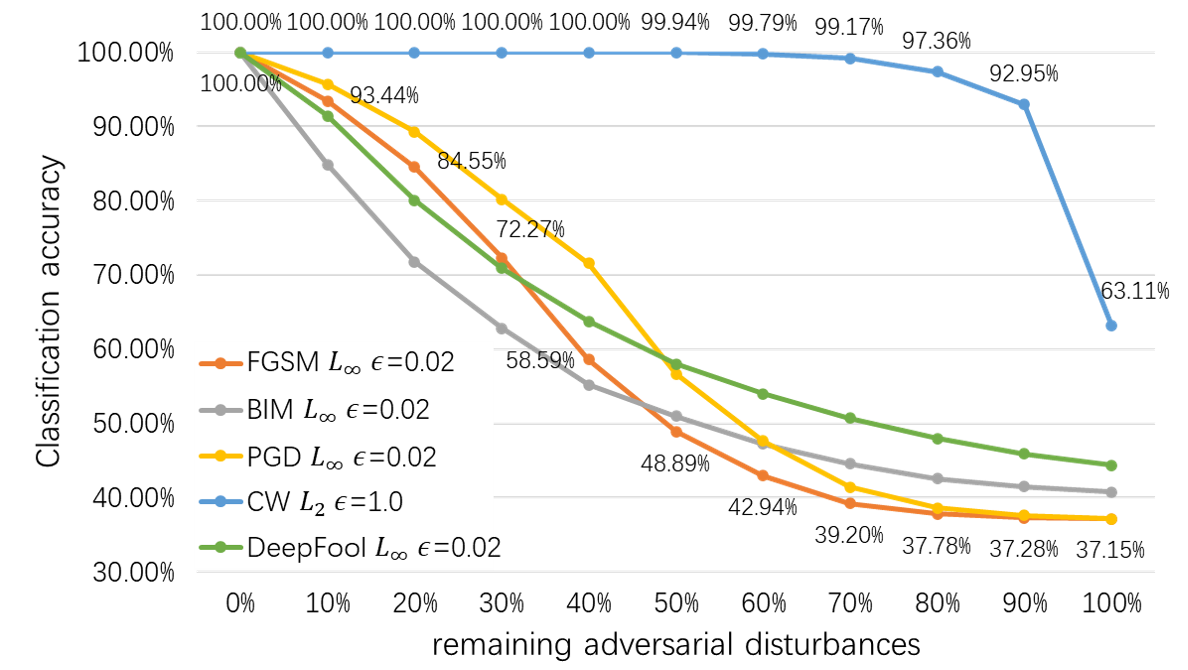}
    \caption{An experiment showing weakened effect of %proportion of the  remaining adversarial disturbances in the original adversarial disturbances 
    partially removed adversarial vectors generated from CIFAR-10 samples on the accuracy of a VGG16 network.} %  classification. Only the values of FGSM attack and CW attack are marked, due to spatial restriction.}
    \label{fig:disturbances}
\end{figure}

\subsection{Mitigating adversarial attacks}
%% explain how the retrained model boosts defense

In our next step, we investigate how the new model trained for masked images can be used to boost defense against adversarial inputs. %Firstly, 
Given an adversarial input $x'=x+\delta$, the new model produces $f_{\theta'}(\tau(x+\delta,c))$. Let $\delta'=\tau(x+\delta, c)-\tau(x,c)$, which is the actual perturbation on the input to $f_{\theta'}$, it is obvious that we have $||\delta'||_p\leq ||\delta||_p$ for any $p$, since only unmasked pixels of $x$ gets affected by $\delta$. %For example, if 
As an example, suppose $3/4$ of the input pixels are %applied with masking, given $c$ randomly chosen, 
masked,
the expected $\ell_1$ norm of $\delta'$ is only about $1/4$ in size of %$||\delta||_1$, 
the $\ell_1$ norm of the original perturbation $\delta$.

In order to measure how much effect of adversarial inputs is weakened by masking, we conduct another preliminary study %to measure such an effect 
by randomly removing part of the perturbation vector $\delta$ from an adversarial input. %(to explain more). 
We train a VGG16 network %is trained on %the train set of 
on CIFAR-10 data set using the conventional method. Then %corresponding 
adversarial samples are generated from clean samples, from which
%by taking normal images %that can be correctly classified by neural network in from the test set. %By comparing the adversarial samples with the original benign images, the adversarial disturbances can be obtained. 
we obtain a perturbation vector by taking the difference between an adversarial input and its corresponding clean image. These vectors are then randomly masked before adding back to the clean images to form a set of weakened attack samples.
%These adversarial disturbances are extracted and randomly masked, that is, part of the disturbances are removed, and then added back to the benign samples. 
The relationship between network accuracy and the proportion of the remaining adversarial disturbance is shown in Figure~\ref{fig:disturbances}, where $100\%$ remaining adversarial disturbance represents strength of the original adversarial inputs, and $20\%$ remaining represents the adversarial inputs with $80\%$ of the perturbation dimensions removed. %in the original adversarial disturbances and the accuracy of network classification after removing part of the adversarial disturbances.
In particular, we find that the CW attack is more sensitive to this perturbation removal operation, %(similar to what is shown in the experiment with the masking operator applied), 
which may be due to that CW generates perturbations %with optimal 
with higher relevance.
%with minimal attack norms. 
For most other attacks, we find that removing at least $60\%$ of the perturbation (i.e., $40\%$ of the attack remaining) is required to deliver significant improvement regarding classification accuracy.
%It can be seen from the experimental results that for CW attacks, as long as a small part of the disturbances are removed, the image classification accuracy can be significantly improved. This also explains why the proposed method can work well on CW attack. For other attack methods, when the masked rate is small, that is, there are more remaining adversarial disturbances, such as $>$ 70\%, the method of masking adversarial disturbances has little improvement on the accuracy of neural network. However, when more adversarial disturbances are masked, such as $>$ 40\% (corresponding, remaining adversarial disturbances $<$ 60\%) the classification accuracy of neural network can be significantly improved with the improvement of mask rate. This shows that it is helpful for neural network to correctly classify images by removing as many adversarial disturbances as possible.

In general, there is often a trade-off between the strength of defense and the degree of accuracy. As shown in Table~\ref{tab:rate} of our experiment in the next section, after fixing a network structure, increasing the percentage of covered pixels (at both the training and test phases) often results in a stronger defense against most of the tested adversarial attacks, but with less accurate classification results on benign (clean) inputs. For example, given the first FGSM attack with perturbation degree $15$ and mask rate of $1/3$ yields a model with defense accuracy of $70.17\%$ on adversarial samples, %(the $1/3$-model), 
which is less effective compared to the $85.3\%$ accuracy when each grid has a $4/5$ probability of being masked. %(the $4/5$-model). 
However, the $4/5$-masked-model only has $76.14\%$ accuracy on benign images, while the $1/3$-masked-model has $85.31\%$ accuracy.

\section{Experiment}
\label{sec:experiment}

The experiments are based on TensorFlow 2.6 and completed on a server running Ubuntu 18.04LTS with a NVIDIA GeForce RTX 3090 GPU. The Adam optimizer is used in training phase with learning rate $0.001$. Foolbox \cite{rauber2017foolbox} is used to generate adversarial samples for the test set. 
We choose three popular DNN models (LeNet with MNIST~\cite{mnist}, VGG16~\cite{VGG} with CIFAR-10~\cite{cifar10} and ResNet18~\cite{ResNet} with SVHN~\cite{svhn}). In order to achieve a classification accuracy close to its corresponding DNN model on benign (clean) images, a MAD model is tested with $20$ randomly masked images transformed from a given image, followed by taking the most favored class of all $20$ tests as the output.

%Chenyi: need a separate paragraph
%\textbf{Note that} 
\paragraph{Basic settings for adversarial attack and defense.} We leave the parameters of all adversarial attack methods as default, %parameters, 
except for the perturbation degree $\epsilon$, which is given along with the ``Attack method'' in the tables. Note that for all experiments in this paper, %corresponding 
adversarial samples are generated only from benign images that are correctly classified by the DNNs to be attacked. 
%Taking the results shown in Table~\ref{tab:effect} for the LeNet MAD model, it has $96.13\%$ accuracy on benign images. 
The adversarial methods also take into account the parameters of the MAD model (in which sense they are white-box attacks). The generated adversarial samples aim to lower the accuracy of the MAD model without masking being applied, e.g., the FGSM L1 with $\epsilon=10$ attack results in a $44.12\%$ accuracy. %of the model, and 
When multiple randomized masking is applied at testing, the accuracy of the  MAD model becomes $68.94\%$ (with $24.82\%$ improvement) facing the same adversarial samples. %via the masking defense).

\begin{figure}[t]
    \centering
    \includegraphics[width=\linewidth]{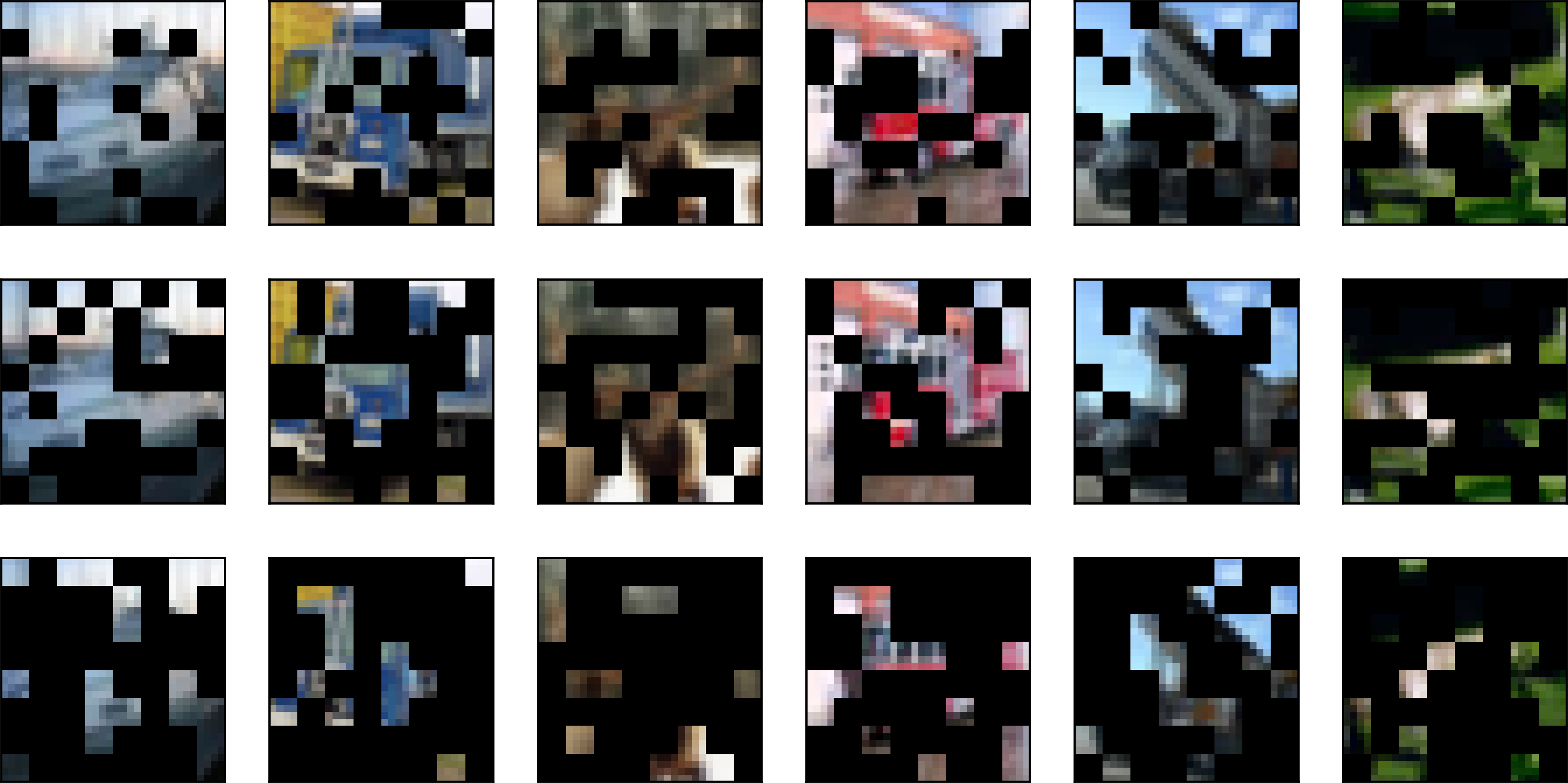}
    %\caption{Masked images with different mask rates. Each column shows how the same image under different mask rates. The mask rates from top to bottom are 1/3, 1/2 and 3/4 respectively.}
    \caption{Images from CIFAR-10 with different masking rates applied: from top to bottom, $1/3$, $1/2$ and $3/4$ rates are respectively applied on the same images.}
    \label{fig:rate}
\end{figure}

\begin{figure}[]
    \centering
    \includegraphics[width=\linewidth]{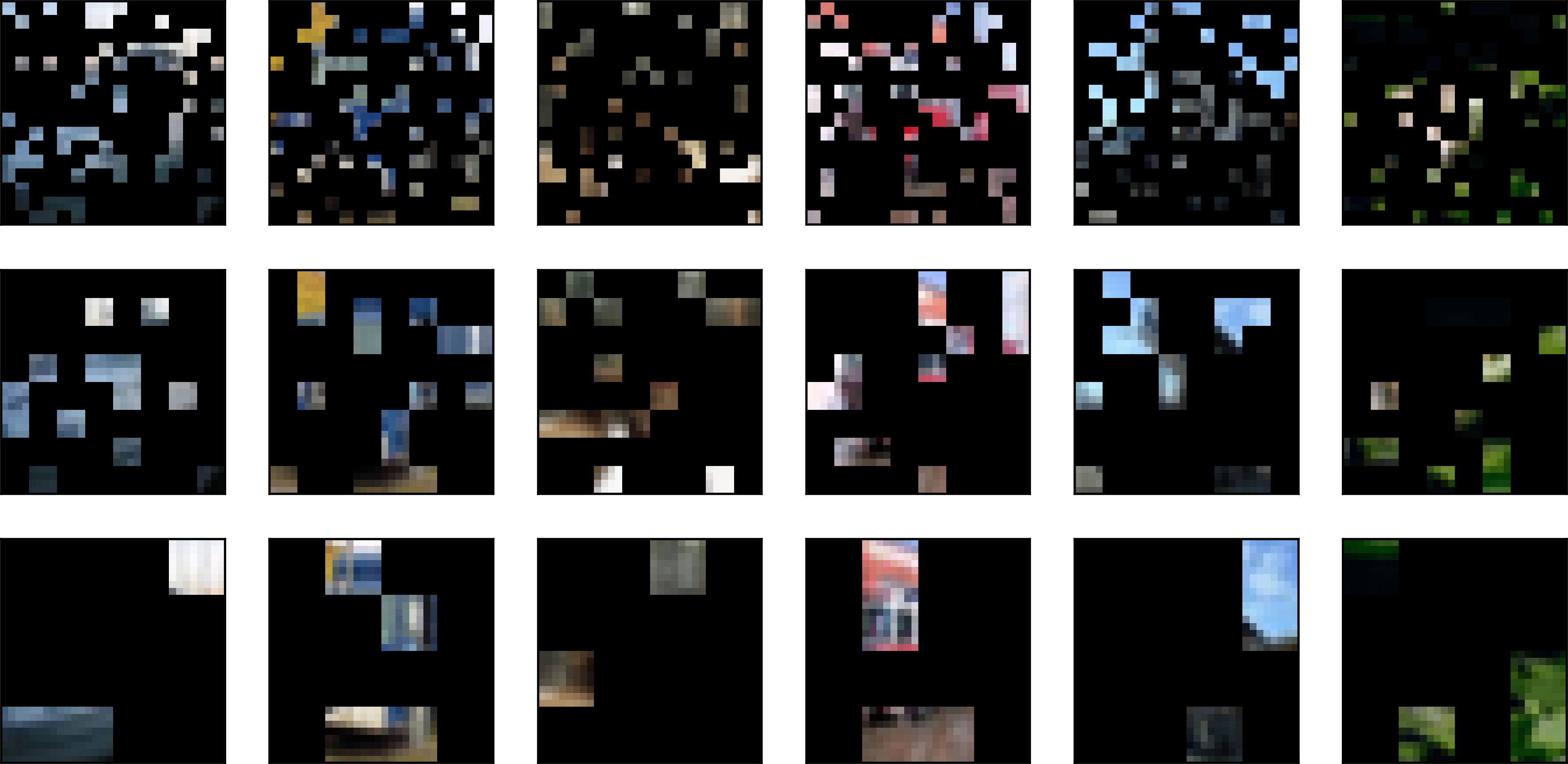}
    %\caption{Masked images with different grid sizes. Each column shows the same image from cifar10 under different grid sizes with the mask rate of 3/4. The grid sizes from top to bottom are 2, 4 and 8 respectively.}
    \caption{$3/4$ masked Images from CIFAR-10 with different grid sizes: from top to bottom,  $2\times 2$, $4\times 4$ and $8\times 8$ grids are used for masking the same images.}
    \label{fig:grid}
\end{figure}

\begin{table}[]
\resizebox{\linewidth}{!}{%
\centering
\begin{tabular}{cccccc}
\hline
Attack method & $0$ & $1/3$ & $1/2$ & $3/4$ & $4/5$ \\ \hline
Benign & \textbf{84.18}\% & 85.31\% & 84.59\% & 82.65\% & 76.14\% \\ \hline
FGSM L1       $\epsilon$=15    & 63.53\% & 70.17\% & 78.37\% & 82.63\% & \textbf{85.30\%} \\
FGSM L1       $\epsilon$=20    & 63.01\% & 64.82\% & 73.70\% & 79.25\% & \textbf{82.14\%} \\
FGSM L2       $\epsilon$=0.3   & 67.36\% & 75.95\% & 82.46\% & 84.62\% & \textbf{87.55\%} \\
FGSM L2       $\epsilon$=0.4   & 66.23\% & 70.18\% & 78.33\% & 81.89\% & \textbf{84.66\%} \\
FGSM Linf     $\epsilon$=0.01  & 48.88\% & 68.87\% & 76.07\% & 81.05\% & \textbf{83.74\%} \\
FGSM Linf     $\epsilon$=0.02  & 40.76\% & 53.77\% & 61.30\% & 68.77\% & \textbf{70.61\%} \\
BIM L1        $\epsilon$=10    & 59.02\% & 67.66\% & 79.63\% & 84.71\% & \textbf{87.88\%} \\
BIM L1        $\epsilon$=15    & 58.23\% & 51.27\% & 68.89\% & 80.91\% & \textbf{84.91\%} \\
BIM L2        $\epsilon$=0.3   & 61.86\% & 61.60\% & 75.69\% & 83.21\% & \textbf{87.04\%} \\
BIM L2        $\epsilon$=0.4   & 61.57\% & 49.33\% & 68.02\% & 80.88\% & \textbf{84.25\%} \\
BIM Linf      $\epsilon$=0.01  & 38.29\% & 57.77\% & 72.21\% & 81.80\% & \textbf{86.14\%} \\
BIM Linf      $\epsilon$=0.015 & 37.15\% & 39.75\% & 58.60\% & 77.06\% & \textbf{82.61\%} \\
PGD L1        $\epsilon$=15    & 58.49\% & 64.60\% & 79.17\% & 85.44\% & \textbf{88.11\%} \\
PGD L1        $\epsilon$=20    & 57.69\% & 54.85\% & 73.67\% & 83.48\% & \textbf{86.89\%} \\
PGD L2        $\epsilon$=0.3   & 62.30\% & 71.87\% & 82.82\% & 86.42\% & \textbf{89.19\%} \\
PGD L2        $\epsilon$=0.4   & 61.63\% & 62.71\% & 77.28\% & 84.45\% & \textbf{87.83\%} \\
PGD Linf      $\epsilon$=0.01  & 40.29\% & 65.37\% & 77.61\% & 84.78\% & \textbf{88.26\%} \\
PGD Linf      $\epsilon$=0.015 & 37.16\% & 49.63\% & 67.64\% & 81.45\% & \textbf{85.63\%} \\
CW L2         $\epsilon$=1     & 63.11\% & 95.46\% & \textbf{95.87\%} & 91.28\% & 93.24\% \\
DeepFool L2   $\epsilon$=0.6   & 53.97\% & 74.01\% & 80.90\% & 84.20\% & \textbf{87.22\%} \\
DeepFool L2   $\epsilon$=0.8   & 52.61\% & 68.15\% & 76.07\% & 81.19\% & \textbf{83.86\%} \\
DeepFool Linf $\epsilon$=0.01  & 50.70\% & 75.72\% & 82.95\% & 85.23\% & \textbf{87.59\%} \\
DeepFool Linf $\epsilon$=0.015 & 44.32\% & 67.31\% & 75.12\% & 80.44\% & \textbf{83.81\%} \\ \hline
\end{tabular}
}
%\caption{Defense effects of different mask rates with the gird size of 8x8 on VGG16 and cifar10. Note that the adversarial attack method works on correctly classified test set samples}
\caption{Defense of different mask rates with the gird size of $8\times 8$ for the training phase and $4\times 4$ for test phase, on a VGG16 model with CIFAR-10 data set}
\label{tab:rate}
\end{table}

%\subsection{On mask rate and grid sizes}
%\label{sec:ex_rate}

\paragraph{On the mask rate and grid size.}
The effectiveness of MAD also depends on two critical hyperparameters. The \textbf{mask rate} is a constant which is used as the probability for a grid to be masked at the training phase as well as test phase. As mentioned %in the previous section
before, a larger mask rate potentially remove more perturbation (which means better defense) at the cost of less usable information for the classifier, which may lead to worse accuracy on clean images. 

%As mentioned in the previous section, a large mask rate will remove more adversarial disturbances and weaken the effect of adversarial attacks. However, under normal circumstances, the adversarial disturbances cannot be separated from the adversarial sample, so the proposed method uses the way of masking both disturbances and images. In this case, a large mask rate will also lead to the loss of more image information, which makes it difficult for neural network classification. 

We carry out an experiment on a selection of mask rates ($1/3$, $1/2$, $3/4$, $4/5$, sample images illustrated in Figure~\ref{fig:rate}) with grid size fixed at $8\times 8$ for the training phase and $4\times 4$ for the test phase, on a VGG16 model with image data from CIFAR-10. Note that this only serves as a rough estimate to the optimal rate from the selected list of values, and the final choice is by no means conclusive. Moreover, it is likely that different DNN models have different optimal mask rates that need to be determined each time before the application of MAD. 
%The effects of different mask rates on classification and adversarial defense are tested on cifar10 data set and VGG16 network. For the convenience and efficiency in achievement, the mask rate in this paper refers to the probability of being masked of each grid, rather than the proportion of the masked part in the whole image. Grid of size 8x8 is used in the training phase and 4x4 in the test phase. For each image correctly classified by neural network in the test set, the corresponding adversarial samples are generated. 
The experimental results are shown in Table~\ref{tab:rate}. 
%% explain why we choose 3/4 ??
It can be seen from the results that higher masking percentage makes better defense but lower accuracy on benign inputs. Considering that in a real life scenario, a DNN may receive a mixed set of benign and adversarial inputs, say $1/2$ benign samples and $1/2$ adversarial samples, for the mask rate of $3/4$ and adversarial samples generated by FGSM with $\epsilon=15$, the real-life defense score can then be calculated as the weighted average $\frac{1}{2}\times 82.63+\frac{1}{2}\times 82.65\% = 82.64\%$ as a fair guess in this hypothetical scenario. %Nevertheless, 
Therefore, we believe that $3/4$ could be a reasonable mask rate which yields acceptable accuracy for both benign and adversarial inputs, and use it for all experiments conducted in rest of the paper.  
%It can be seen from the experimental results that more mask of training images will slightly reduce the classification accuracy of neural network when facing benign samples due to the loss of image information. Smaller mask rate will even improve the classification accuracy of neural network for benign samples. This may be due to the reduction of over fitting of neural network by masking operation. As mentioned in Section \ref{sec:ex_disturbances}, with the increase of mask rate, the effect of adversarial defense also increases. From the experimental results, in most of the adversarial attack methods, the improvement of adversarial defense effect when the mask rate is 4/5 is limited than that when the mask rate is 3/4. At the same time, the increment of mask rate will reduce the classification accuracy of benign samples. Therefore, the proposed method uses a mask rate of 3/4.

\begin{table}[] \scriptsize
\resizebox{\linewidth}{!}{
\centering
\begin{tabular}{ccccc}
\hline
 & Attack method & Attack & Defense & Improvement \\ \hline
\multirow{24}{*}{\rotatebox{90}{LeNet + MNIST}}
 & Benign & & 96.13\% & \\
 & FGSM L1         $\epsilon$=10    & 44.12\% & 68.94\% & 24.82\% \\
 & FGSM L1         $\epsilon$=20    & 28.35\% & 49.50\% & 21.15\% \\
 & FGSM L2         $\epsilon$=0.8   & 34.85\% & 58.60\% & 23.75\% \\
 & FGSM L2         $\epsilon$=1.0   & 30.28\% & 52.82\% & 22.54\% \\
 & FGSM Linf       $\epsilon$=0.03  & 36.26\% & 61.82\% & 25.56\% \\
 & FGSM Linf       $\epsilon$=0.04  & 28.58\% & 50.31\% & 21.73\% \\
 & BIM L1          $\epsilon$=4     &  8.98\% & 59.49\% & 50.51\% \\
 & BIM L1          $\epsilon$=6     &  2.26\% & 44.17\% & 41.91\% \\
 & BIM L2          $\epsilon$=0.2   & 16.55\% & 69.23\% & 52.68\% \\
 & BIM L2          $\epsilon$=0.3   &  3.79\% & 49.04\% & 45.25\% \\
 & BIM Linf        $\epsilon$=0.015 & 11.72\% & 70.94\% & 59.22\% \\
 & BIM Linf        $\epsilon$=0.02  &  3.62\% & 56.33\% & 52.71\% \\
 & PGD L1          $\epsilon$=4     & 16.04\% & 71.79\% & 55.75\% \\
 & PGD L1          $\epsilon$=6     &  3.37\% & 54.75\% & 51.38\% \\
 & PGD L2          $\epsilon$=0.2   & 22.98\% & 77.87\% & 54.89\% \\
 & PGD L2          $\epsilon$=0.4   &  1.26\% & 48.85\% & 47.59\% \\
 & PGD Linf        $\epsilon$=0.015 & 11.64\% & 70.90\% & 59.26\% \\
 & PGD Linf        $\epsilon$=0.02  &  3.27\% & 56.38\% & 53.11\% \\
 & CW L2           $\epsilon$=1.5   &  0.20\% & 93.39\% & 93.19\% \\
 & DeepFool L2     $\epsilon$=0.6   & 32.62\% & 61.05\% & 28.43\% \\
 & DeepFool L2     $\epsilon$=0.8   & 26.49\% & 51.39\% & 24.90\% \\
 & DeepFool Linf   $\epsilon$=0.05  & 24.77\% & 47.26\% & 22.49\% \\
 & DeepFool Linf   $\epsilon$=0.07  & 22.50\% & 39.76\% & 17.26\% \\ \hline
\multirow{24}{*}{\rotatebox{90}{VGG16 + CIFAR-10}}
 & Benign & & 82.65\% & \\
 & FGSM L1         $\epsilon$=15    & 35.48\% & 82.63\% & 47.15\% \\
 & FGSM L1         $\epsilon$=20    & 33.02\% & 79.25\% & 46.23\% \\
 & FGSM L2         $\epsilon$=0.3   & 38.29\% & 84.62\% & 46.33\% \\
 & FGSM L2         $\epsilon$=0.4   & 35.27\% & 81.89\% & 46.62\% \\
 & FGSM Linf       $\epsilon$=0.01  & 33.02\% & 81.05\% & 48.03\% \\
 & FGSM Linf       $\epsilon$=0.02  & 28.27\% & 68.77\% & 40.50\% \\
 & BIM L1          $\epsilon$=10    & 10.23\% & 84.71\% & 74.48\% \\
 & BIM L1          $\epsilon$=15    &  6.70\% & 80.91\% & 74.21\% \\
 & BIM L2          $\epsilon$=0.3   &  9.43\% & 83.21\% & 73.78\% \\
 & BIM L2          $\epsilon$=0.4   &  8.06\% & 80.88\% & 72.82\% \\
 & BIM Linf        $\epsilon$=0.01  &  3.16\% & 81.80\% & 78.64\% \\
 & BIM Linf        $\epsilon$=0.015 &  1.06\% & 77.06\% & 76.00\% \\ 
 & PGD L1          $\epsilon$=15    &  8.75\% & 85.44\% & 76.69\% \\
 & PGD L1          $\epsilon$=20    &  6.36\% & 83.48\% & 77.12\% \\
 & PGD L2          $\epsilon$=0.3   & 12.47\% & 86.42\% & 73.95\% \\
 & PGD L2          $\epsilon$=0.4   &  9.01\% & 84.45\% & 75.44\% \\
 & PGD Linf        $\epsilon$=0.01  &  6.13\% & 84.78\% & 78.65\% \\
 & PGD Linf        $\epsilon$=0.015 &  2.30\% & 81.45\% & 79.15\% \\ 
 & CW L2           $\epsilon$=1     & 15.89\% & 91.28\% & 75.39\% \\
 & DeepFool L2     $\epsilon$=0.6   & 34.64\% & 84.20\% & 49.56\% \\
 & DeepFool L2     $\epsilon$=0.8   & 31.81\% & 81.19\% & 49.38\% \\
 & DeepFool Linf   $\epsilon$=0.01  & 35.72\% & 85.23\% & 49.51\% \\
 & DeepFool Linf   $\epsilon$=0.015 & 32.26\% & 80.44\% & 48.18\% \\ \hline 
\multirow{24}{*}{\rotatebox{90}{ResNet18 + SVHN}}
& Benign & & 91.65\% & \\
& FGSM L1         $\epsilon$=15 	  & 50.47\% & 76.84\% & 26.37\% \\
& FGSM L1         $\epsilon$=20	  & 45.85\% & 70.25\% & 24.40\% \\
& FGSM L2         $\epsilon$=0.4	  & 53.17\% & 80.40\% & 27.23\% \\
& FGSM L2         $\epsilon$=0.6	  & 46.77\% & 71.48\% & 24.71\% \\
& FGSM Linf       $\epsilon$=0.01	& 57.12\% & 85.40\% & 28.28\% \\
& FGSM Linf       $\epsilon$=0.02	& 46.06\% & 68.95\% & 22.89\% \\
& BIM L1          $\epsilon$=10	  & 15.46\% & 80.37\% & 64.91\% \\
& BIM L1          $\epsilon$=12	  & 11.54\% & 75.85\% & 64.31\% \\
& BIM L2          $\epsilon$=0.25	& 21.75\% & 84.40\% & 62.65\% \\
& BIM L2          $\epsilon$=0.3	  & 16.59\% & 81.37\% & 64.78\% \\
& BIM Linf        $\epsilon$=0.01	& 16.06\% & 86.31\% & 70.25\% \\
& BIM Linf        $\epsilon$=0.02	&  2.58\% & 71.00\% & 68.42\% \\
& PGD L1          $\epsilon$=14    & 15.09\% & 84.14\% & 69.05\% \\
& PGD L1          $\epsilon$=16    & 12.11\% & 82.35\% & 70.24\% \\
& PGD L2          $\epsilon$=0.4   & 16.40\% & 84.22\% & 67.82\% \\
& PGD L2          $\epsilon$=0.5   & 12.17\% & 81.37\% & 69.20\% \\
& PGD Linf        $\epsilon$=0.015 & 10.10\% & 84.45\% & 74.35\% \\
& PGD Linf        $\epsilon$=0.02  &  4.69\% & 80.51\% & 75.82\% \\
& CW L2           $\epsilon$=1.8   & 21.70\% & 96.05\% & 74.35\% \\
& DeepFool L2     $\epsilon$=0.5   & 59.78\% & 87.09\% & 27.31\% \\
& DeepFool L2     $\epsilon$=0.7   & 53.87\% & 82.22\% & 28.35\% \\
& DeepFool Linf   $\epsilon$=0.01	& 57.10\% & 88.06\% & 30.96\% \\
& DeepFool Linf   $\epsilon$=0.02	& 46.39\% & 72.98\% & 26.59\% \\ \hline
\end{tabular}
}
%\caption{The effect of MAD on classification accuracy}
\caption{Defense against adversarial attacks by MAD}
\label{tab:effect}
\end{table}

%\subsection{Grid size}\label{sec:grid}

The \textbf{grid size} is another hyperparameter to be determined before the main experiment. Intuitively, a larger grid size allows better continuity of information on average that is preserved in a masked image, examples illustrated in Figure~\ref{fig:grid}.
%
%The size of the grid determines the continuity of information in the masked images. Figure \ref{fig:grid} shows some masked images with different grid sizes. Too small grid size will reduce the continuity of information and affect the classification effect of neural network. For example, when using grid size of 2x2, the classification accuracy of VGG16 on cifar10's test set is only 61.13\%. The large grid size makes the retained information more continuous, but it also increases the difficulty of restoring the masked part of the information.
%
Table~\ref{tab:grid} presents the preliminary experimental results on a VGG model with CIFAR-10 data set for a selection of grid size values in the masking operation at the time of training and testing, respectively. Given images in CIFAR-10 are of size $32\times 32$, we try a number of combinations for values that divide $32$, and again, we do not claim that we are able to find an optimal setting in such an experiment. If we focus on the three columns where $8\times 8$ grids are used for training, the results seem to suggest that using a larger grid in the test phase produces a better classification accuracy on benign images, and a smaller grid produces a better defense. This time, we choose  $8\times 8$ (training) and $4\times 4$ (testing) for the VGG16 model and CIFAR-10 data set, and adopt this same setting for ResNet and SVHN that also use $32\times 32$ images. For the LeNet model and MNIST data set ($28\times 28$ images), we use $7\times 7$ grids for both training and testing. Likewise, for DNN models not considered in this paper, the grid size to be applied may be different.
%Considering the size of the images (32x32), grid sizes of 4x4 and 8x8 are tested. As the results show, using a large grid in the training phase is helpful for the network to better classify benign images. In the test phase, using a smaller grid size can obtain better adversarial defense effect, but it will reduce the classification accuracy of benign samples. After weighing the classification accuracy of benign samples and the effect of adversarial defense, the proposed method uses 7x7 (for MNIST) or 8x8 (for cifar10 and SVHN) grids in the training phase and 4x4 grids in the test phase.

\begin{table}[] %\scriptsize
\centering
\resizebox{\linewidth}{!}{
\begin{tabular}{cccccc}
\hline
Training & 4$\times$4 & 4$\times$4 & 8$\times$8 & 8$\times$8 & 8$\times$8 \\
Test & 2$\times$2 & 4$\times$4 & 2$\times$2 & 4$\times$4 & 8$\times$8 \\ \hline
Benign & 73.23\% & 76.92\% & 63.11\% & 82.65\% & 82.89\% \\ \hline
FGSM L1       $\epsilon$=15    & 86.63\%  & 83.84\% & 85.34\% & 82.63\% & 75.68\% \\
FGSM L1       $\epsilon$=20    & 83.20\%  & 79.39\% & 82.74\% & 79.25\% & 71.72\% \\
FGSM L2       $\epsilon$=0.3   & 89.54\%  & 87.13\% & 86.17\% & 84.62\% & 79.84\% \\
FGSM L2       $\epsilon$=0.4   & 86.63\%  & 83.41\% & 85.34\% & 81.89\% & 75.11\% \\
FGSM Linf     $\epsilon$=0.01  & 86.10\%  & 82.11\% & 84.42\% & 81.05\% & 73.63\% \\
FGSM Linf     $\epsilon$=0.02  & 73.64\%  & 67.07\% & 73.93\% & 68.77\% & 59.87\% \\
BIM L1        $\epsilon$=10    & 89.21\%  & 85.66\% & 87.64\% & 84.71\% & 75.67\% \\
BIM L1        $\epsilon$=15    & 84.62\%  & 79.36\% & 85.61\% & 80.91\% & 65.50\% \\
BIM L2        $\epsilon$=0.3   & 87.52\%  & 83.55\% & 86.12\% & 83.21\% & 71.88\% \\
BIM L2        $\epsilon$=0.4   & 83.64\%  & 78.06\% & 85.50\% & 80.88\% & 65.82\% \\
BIM Linf      $\epsilon$=0.01  & 86.60\%  & 81.86\% & 86.47\% & 81.80\% & 68.67\% \\
BIM Linf      $\epsilon$=0.015 & 81.26\%  & 72.97\% & 83.62\% & 77.06\% & 56.05\% \\
PGD L1        $\epsilon$=15    & 88.99\%  & 84.76\% & 86.64\% & 85.44\% & 76.14\% \\
PGD L1        $\epsilon$=20    & 85.78\%  & 80.08\% & 86.23\% & 83.48\% & 71.13\% \\
PGD L2        $\epsilon$=0.3   & 90.35\%  & 87.03\% & 88.07\% & 86.42\% & 78.77\% \\
PGD L2        $\epsilon$=0.4   & 88.01\%  & 83.48\% & 87.20\% & 84.45\% & 74.79\% \\
PGD Linf      $\epsilon$=0.01  & 88.64\%  & 84.05\% & 86.61\% & 84.78\% & 73.89\% \\
PGD Linf      $\epsilon$=0.015 & 84.57\%  & 77.20\% & 84.46\% & 81.45\% & 65.54\% \\
CW L2         $\epsilon$=1     & 94.78\%  & 95.25\% & 89.53\% & 91.28\% & 93.94\% \\
DeepFool L2   $\epsilon$=0.6   & 88.54\%  & 87.10\% & 85.03\% & 84.20\% & 80.31\% \\
DeepFool L2   $\epsilon$=0.8   & 85.73\%  & 82.75\% & 82.65\% & 81.19\% & 75.53\% \\
DeepFool Linf $\epsilon$=0.01  & 90.33\%  & 88.21\% & 85.42\% & 85.23\% & 80.42\% \\
DeepFool Linf $\epsilon$=0.015 & 86.03\%  & 82.66\% & 82.16\% & 80.44\% & 74.92\% \\ \hline
\end{tabular}%
}
\caption{Defense effects of different grid sizes with mask rate of 3/4 on VGG16 and CIFAR-10.}
\label{tab:grid}
\end{table}

%\subsection{Comparison}
\subsection*{The main result and comparison}
\label{sec:comparison}

The main experiment is conducted on three different models, LeNet with MNIST data set, VGG16 with CIFAR-10, and ResNet18 with SVHN, with the results presented in Table~\ref{tab:effect}. 
Of all the attacks, CW~\cite{Carlini2017Towards} is the most successfully defended method by MAD. A possible explanation is that CW usually produces highly correlated perturbation, which can be more easily countered by masking as presented in Figure~\ref{fig:disturbances}. For each attack method with different perturbation degrees ($\epsilon$), MAD tends to achieve a better defense on the attack with a smaller $\epsilon$, which is not surprising.

We compare MAD with four state-of-the-art adversarial defence methods, including MagNet~\cite{meng2017magnet}, Denoised smoothing~\cite{salman2020denoised} \footnote{https://github.com/microsoft/denoised-smoothing}, Parseval networks~\cite{pmlr-v70-cisse17a} \footnote{https://github.com/mathialo/parsnet}, and PCL~\cite{mustafa2019adversarial} on the same backbone VGG16 model with default setting, and all tested with adversarial samples that are generated from CIFAR-10. 
%%This section compares the proposed method with several other adversarial defense methods. The backbone network in these methods is modified to VGG16 and use the parameters given in the code. 
Note that the adversarial samples used for comparison are generated only from correctly classified benign samples. %, with the white-box attack mode. %To be fair, 
None of these defense methods uses adversarial samples in the training phase. %The comparison in cifar10 and VGG16 between these methods and the proposed methods is shown in Table \ref{tab:comparison}. 
The results are shown in Table~\ref{tab:comparison}. 

%To reach a more conclusive 
In order to make the comparison more comprehensive and more convincing, the %competing 
selected defense methods %are chosen from different classes of defense and with 
have inherently different underlying approaches. 
MagNet and Denoised smoothing apply additional denoising structures to reduce the adversarial perturbations. Parseval network restricts the layers' Lipschitz constant to be less than $1$ during the training. PCL enforces hidden layer features of different classes to be apart as much as possible during the training, which is imposed by adding new branch structures and introducing a new loss function. %, in order to increase the level of difficulty for adversarial attacks.
For our experiment on the comparison, the
implementations of Denoised smoothing and PCL are publicly available from their corresponding authors, and the codes for MagNet and Parseval network are available from (third party) Gokul Karthik~\footnote{https://github.com/GokulKarthik/MagNet.pytorch} and standard Python library \emph{parsnet}\footnote{https://github.com/mathialo/parsnet}, respectively.
%and the code for Parseval network is taken from the standard Python library \emph{parsnet} \footnote{https://github.com/mathialo/parsnet}.
%Implementations of three (out of four) attack methods are publicly available from their corresponding authors, and the code for Parseval network is taken from the standard Python library \emph{parsnet} \footnote{https://github.com/mathialo/parsnet}.

From the results in Table~\ref{tab:comparison}, all methods have an acceptable %a close %to optimal 
accuracy on benign inputs. %(compared to the $84.18\%$ accuracy of the vanilla VGG16). 
However, regarding the list of tested adversarial attacks, MAD outperforms %other
these state-of-the-art defense approaches in most cases.

%MagNet \cite{meng2017magnet} \footnote{https://github.com/GokulKarthik/MagNet.pytorch} and Denoised smoothing\cite{salman2020denoised} \footnote{https://github.com/microsoft/denoised-smoothing} reduce the adversarial disturbance in the image input to the DNN by adding a denoising network before it. The vanilla VGG16 can achieve 84.18\% classification accuracy on the test set without the denoising network. Both denoising networks reduce the classification accuracy of benign samples, but the classification accuracy of adversarial samples after denoising is improved. And Denoised smoothing achieve the best defense effect on CW attack.

%Parseval networks \cite{pmlr-v70-cisse17a} limit the layers' Lipschitz constant to less than 1. Because it does not publish the source code, the implementation of python library \emph{parsnet} \footnote{https://github.com/mathialo/parsnet} is used. And only the parameters of the convolution layers in the network are limited. PCL \cite{mustafa2019adversarial} \footnote{https://github.com/aamir-mustafa/pcl-adversarial-defense} make the hidden layer features of different classes of images separated as far as possible, by adding some branch structures to the network and designing the loss function to increase the adversarial attack difficulty and reduce the attack effect.

%As shown in Table \ref{tab:comparison}, when considering both the classification accuracy of benign samples and adversarial defense effect , the performance of the proposed method is better than that of other methods in most of the adversarial attack methods.

\begin{table}[]
\newcommand{\tabincell}[2]{\begin{tabular}{@{}#1@{}}#2\end{tabular}}
\centering
\resizebox{\linewidth}{!}{
\begin{tabular}{crrrrr}
\hline
Attack method & \multicolumn{1}{c}{MagNet} & \tabincell{c}{Denoised\\smoothing} & \tabincell{c}{Parseval\\networks} & \multicolumn{1}{c}{PCL} & \multicolumn{1}{c}{MAD} \\ \hline
Benign & 82.32\% & 77.92\% & 77.89\% & \textbf{83.47\%} & 82.65\% \\ \hline
FGSM L1       $\epsilon$=15    & 67.41\%            & 81.97\%           & 43.20\% & 74.94\% & \textbf{82.63\%} \\
FGSM L1       $\epsilon$=20    & 65.68\%            & 76.50\%           & 33.11\% & 70.53\% & \textbf{79.25\%} \\
FGSM L2       $\epsilon$=0.3   & 72.15\%            & \textbf{85.88\%}  & 52.41\% & 79.54\% & 84.62\% \\
FGSM L2       $\epsilon$=0.4   & 69.75\%            & 81.70\%           & 42.69\% & 75.32\% & \textbf{81.89\%} \\
FGSM Linf     $\epsilon$=0.01  & 57.81\%            & \textbf{82.83\%}  & 44.22\% & 75.56\% & 81.05\% \\
FGSM Linf     $\epsilon$=0.02  & 44.96\%            & 65.25\%           & 20.75\% & 62.72\% & \textbf{68.77\%} \\
BIM L1        $\epsilon$=10    & 66.56\%            & 80.11\%           & 45.23\% & 64.03\% & \textbf{84.71\%} \\
BIM L1        $\epsilon$=15    & 62.63\%            & 68.58\%           & 26.22\% & 48.03\% & \textbf{80.91\%} \\
BIM L2        $\epsilon$=0.3   & 67.63\%            & 77.36\%           & 38.57\% & 61.26\% & \textbf{83.21\%} \\
BIM L2        $\epsilon$=0.4   & 65.73\%            & 68.06\%           & 24.96\% & 48.34\% & \textbf{80.88\%} \\
BIM Linf      $\epsilon$=0.01  & 51.34\%            & 75.51\%           & 29.54\% & 59.83\% & \textbf{81.80\%} \\
BIM Linf      $\epsilon$=0.015 & 42.84\%            & 62.41\%           & 13.52\% & 43.31\% & \textbf{77.06\%} \\
PGD L1        $\epsilon$=15    & 67.08\%            & 79.83\%           & 37.84\% & 66.63\% & \textbf{85.44\%} \\
PGD L1        $\epsilon$=20    & 64.66\%            & 72.81\%           & 24.64\% & 53.18\% & \textbf{83.48\%} \\
PGD L2        $\epsilon$=0.3   & 71.98\%            & 84.65\%           & 48.20\% & 76.08\% & \textbf{86.42\%} \\
PGD L2        $\epsilon$=0.4   & 69.44\%            & 79.17\%           & 35.14\% & 66.70\% & \textbf{84.45\%} \\
PGD Linf      $\epsilon$=0.01  & 56.92\%            & 81.17\%           & 36.89\% & 72.22\% & \textbf{84.78\%} \\
PGD Linf      $\epsilon$=0.015 & 47.79\%            & 70.74\%           & 19.35\% & 54.74\% & \textbf{81.45\%} \\
CW L2         $\epsilon$=1     & 92.58\%   & \textbf{99.74\%}              & 43.42\% & 89.48\% &         91.28\% \\
DeepFool L2   $\epsilon$=0.6   & 63.84\%            & 65.04\%           & 34.10\% & 49.95\% & \textbf{84.20\%} \\
DeepFool L2   $\epsilon$=0.8   & 60.45\%            & 57.97\%           & 44.51\% & 47.68\% & \textbf{81.19\%} \\
DeepFool Linf $\epsilon$=0.01  & 62.79\%            & 81.43\%           & 30.49\% & 58.80\% & \textbf{85.23\%} \\
DeepFool Linf $\epsilon$=0.015 & 54.26\%            & 71.68\%           &  9.49\% & 53.28\% & \textbf{80.44\%} \\ \hline
\end{tabular}
}
\caption{Comparison of effects of different adversarial defense methods on VGG16 and CIFAR-10.}
\label{tab:comparison}
\end{table}

% \subsection{Training detail}
% \label{sec:detail}

% Firstly, the network is trained on the training set of the complete data set, which is the same as the traditional training method. Then, the masked images generated by the method mentioned in Section \ref{sec:method} are used for training. The training uses Adam optimizer, with the learning rate of 0.001, a total of 300 epochs.

% Deepfool \cite{rauber2017foolbox} is used to generate corresponding adversarial sample for each image correctly classified in the test set. Parameters of adversarial attack methods are set as the default parameters, except the perturbation degree $\epsilon$.

%% white-box adaptive attack
% need some discussion later
%\subsection*{Adaptive White-box Attacks}

\section{Conclusion}
\label{sec:conclusion}
In this paper, we have proposed a new mask-based adversarial defense method called MAD, and have conducted extensive experiments showing that our method provides an effective defense against a variety of adversarial attacks. %Compared with other adversarial defense methods, MAD does not need additional denoising structure, or any change to the existing DNN, but still achieves state-of-the-art defense.

Moreover, since the  attack algorithms are allowed to have access to the parameters of MAD, our defense seems to provide a way to withstand adaptive white-box attacks, as the randomized masking technique may enforce attackers to consider an attack space of size exponential to the number of grids required to cover the input. %, which can be too large in a real world scenario.
We plan to study detailed measurement of this effect on adaptive white-box adversarial attack methods in our future work.

%This paper proposes a mask-based adversarial defense training method. According to the principle  of adversarial attacks and experimental results, reducing adversarial disturbances helps the neural network to correctly classify adversarial examples. In the proposed method, parts of the images are masked to reduce the adversarial disturbance. The masked images are used in the training and test phases. Compared with other adversarial defense methods, the proposed method does not need additional network structure or loss function. Experimental results on different data sets and networks show that the proposed method can effectively improve the adversarial defense ability of the network.


\newpage

\bibliographystyle{named}
\bibliography{cas-refs}
%\bibliographystyle{elsarticle-num} 





% \end{thebibliography}
\end{document}